\documentclass[runningheads]{llncs}

\usepackage[T1]{fontenc} 
\usepackage{graphicx}
\usepackage{float}
\usepackage[table,xcdraw]{xcolor}
\usepackage{hyperref}
\usepackage{tabularx}
\usepackage{booktabs}
\usepackage{icomma} 
\usepackage{subcaption}

\begin{document}

\title{Development of Hybrid Artificial Intelligence Training on Real and Synthetic Data}
\subtitle{Benchmark on Two Mixed Training Strategies}
\titlerunning{Development of Hybrid AI Training on Real and Synthetic Data}
\author{Paul Wachter \inst{1} \orcidID{0000-0002-6224-6140} \and
Lukas Niehaus \inst{2} \orcidID{0009-0009-9978-6851} \and
Julius Schöning \inst{1} \orcidID{0000-0003-4921-5179}
}

\authorrunning{P. Wachter et al.}

\institute{Faculty of Engineering and Computer Science, Osnabrück University of Applied Sciences, Osnabrueck, Germany \\
\email{\{p.wachter,j.schoening\}@hs-osnabrueck.de} \and
Institute of Cognitive Science, Osnabrück University, Osnabrueck, Germany\\
\email{luniehaus@uni-osnabrueck.de}
}

\maketitle              

\begin{abstract}
\setcounter{footnote}{0}% Reset footnote counter
\renewcommand{\thefootnote}{\alph{footnote}}
Synthetic data has emerged as a cost-effective alternative to real data for training artificial neural networks (ANN). 
However, the disparity between synthetic and real data results in a \emph{domain gap}. That gap leads to poor performance and generalization of the trained ANN when applied to real-world scenarios.
Several strategies have been developed to bridge this gap, which combine synthetic and real data, known as mixed training using hybrid datasets.
While these strategies have been shown to mitigate the domain gap, a systematic evaluation of their generalizability and robustness across various tasks and architectures remains underexplored. 
To address this challenge, our study comprehensively analyzes two widely used mixing strategies on three prevalent architectures and three distinct hybrid datasets.
From these datasets, we sample subsets with varying proportions of synthetic to real data to investigate the impact of synthetic and real components.
The findings of this paper provide valuable insights into optimizing the use of synthetic data in the training process of any ANN, contributing to enhancing robustness and efficacy cf. \footnote{Supplementary code and results: https://hs-osnabrueck.de/prof-dr-julius-schoening/ki2025}\label{footnote_code}

\end{abstract}

\section{Introduction}
Synthetic data has recently gained considerable attention as a cost-effective way to generate training data~\cite{Niko} for artificial neural networks (ANN). Synthetic data, defined as information not derived from real-world sources, 
inherently lacks realism. This absence creates a discrepancy between 
synthetic and real examples, which can be termed the \emph{reality gap}~\cite{8575297}. The reality gap is a subset of the broader \emph{domain gap}, which limits
 the exclusive use of synthetic data in ANN training since the target 
domain usually is the real-world. 
As a result, applying ANNs, which are solely trained on synthetic data, to real-world scenarios requires domain adaptation to transfer knowledge from one domain to the other~\cite{Peng_2018_CVPR_Workshops}. 

One key area of domain adaptation research focuses on enhancing the 
quality and diversity of synthetic data to bridge the domain gap \cite{Niko}. 
This can be achieved by increasing the realism of synthetic data either 
during its generation or through post-processing techniques. 
Enhancing realism during generation requires an advanced simulation framework that accurately models real-world conditions~\cite{greenmatter}. 
Such frameworks often prove costly and labor-intensive~\cite{9954643}. Alternatively, image transformation 
methods, such as neural style transfer, improve realism in post-processing~\cite{ettedgui2022procstboostingsemanticsegmentation}, 
although these methods often suffer from training instabilities and may introduce artifacts~\cite{Zhu2017UnpairedIT}.

Another research focus targets the architecture of ANNs, aiming to enable them to learn domain-invariant features, thereby reducing the domain gap~\cite{Wang2018DeepVD}. 
However, designing such architectures is inherently challenging because they must effectively abstract away domain-specific details to maintain generality. 
Moreover, many of these architectures are developed for multi-domain adaptation~\cite{peng2019moment}, including domains with no available data. 
These assumptions may limit their effectiveness in addressing the specific reality gap, where real data is usually accessible.

Consequently, this paper examines a mixed training approach for domain adaptation, which integrates synthetic and real data into a unified hybrid dataset. By integrating both types into a hybrid set, mixed training 
allows models to learn features from synthetic part, while real-world 
data introduces authentic visual and contextual variations, thereby mitigating the lack of realism in the synthetic data and
bridging the domain gap. This approach can be used with every ANN architecture, thus making it broadly applicable, while omitting the need for highly specialized ANNs when relaying on synthetic sources. 
Mixed training prominently follows two strategies: \emph{simple mixed} (SM), 
where both data types are used simultaneously, and sequentially 
\emph{fine-tuned} (FT), in which ANNs are pretrained on synthetic data and 
afterwards fine-tuned with real data. Although these methods are widely applied, their underlying mechanisms 
and performance under varying conditions remains only partially 
understood.

In response to these limitations, this study systematically evaluates both mixed training 
strategies across three structurally distinct datasets, varying ratios 
of synthetic-to-real data and three different ANN architectures on image classification tasks. Unlike
 previous studies that boost model performance by expanding the hybrid 
dataset with additional synthetic examples, our approach maintains a 
constant dataset size while gradually increasing the synthetic-to-real proportion. Thus, our goal is not limited to improving the overall performance, but to conduct a precise assessment of each 
strategies efficacy under varying conditions. This design offers actionable insights to guide the practical application of synthetic data in real-world settings.

\section{Related Work}

Although many publications have successfully applied mixed training to 
reduce the domain gap~\cite{KIM2023104771,app15010354,7780721,rajpura2017objectdetectionusingdeep}, most studies concentrate on the generation of synthetic data rather than examining the mixed training strategies themselves \cite{article11,Wachter2024}.

Nowruzi et al.~\cite{nowruzi_how_2019} evaluated an object detector for 
cars and pedestrians using three synthetic and three real datasets. They
 created hybrid training sets by combining synthetic data
with four different, low ratios of real data. 
Moreover, they compared the SM strategy
 with the FT approach for their tasks and concluded that FT more effectively reduces the domain gap. However, the study did not consider the effects of varying dataset sizes
and class distributions. In contrast, Burdorf et al.~\cite{burdorf2022reducingrealworlddata,9607667} did not confirm the advantages of FT for a very similar task as reported by Nowruzi et al.~\cite{nowruzi_how_2019}.

In \cite{Vanherle_2022_BMVC}, Vanherle et al. compared the SM and FT strategies across subsets with the same overall size but different 
synthetic-to-real ratios; notably, the subsets were imbalanced. They employ the DIMO dataset~\cite{de2022dataset}, which contains real and synthetic images, to train and test a Mask-Region Based Convolutional Neural Networks (Mask R-CNN) for object detection. Although the FT strategy yielded better 
performance, the study involved pretraining on the COCO dataset before 
introducing synthetic examples. Moreover, synthetic data was used solely to 
train the heads of the pretrained model. 
These methodological choices impact network performance~\cite{8981600} and may interact with the mixed training strategies, potentially confounding true differences.

Together, these studies leave several questions unresolved. Reported gains could be due to uncontrolled factors, like dataset size, class imbalance, or pre-raining on large scale datasets with freezing of layers, that obscure the true effect of the mixing strategy. Most experiments rely on a single hybrid dataset and one ANN, making it unclear whether conclusions generalize across synthetic data types or ANN architectures. A more controlled, cross-dataset, cross-architecture analysis is therefore required, which we address in this study.

\section{Methodology}
The complete code, configurations and numerical results of our study can be found at~\ref{footnote_code}. In general, mixed training using real and synthetic data can be applied to any neural network and hybrid dataset, regardless of the architecture, synthetic data type or synthetic-to-real ratio. Consequently, this study covers a broad range of possible configurations of these key factors to evaluate their influence on the two training strategies SM and FT.
Our study is designed to expose and isolate the effects on model performance, instead of aiming for maximum accuracy in each configuration.
Accordingly, we minimize other factors which can influence the training process, such as advanced optimization algorithms, regularization heuristics, data augmentation or hyperparameter tuning.
In the following, we lay out the choices of our ANN architectures, hybrid datasets and synthetic-to-real proportions, as well as the implementation of the SM and FT training strategies.

\subsection{Application Task}
We choose image classification as the application task of our study. Image classification is considered one of the most fundamental tasks within the domain of AI-based and classical computer vision~\cite{schmidhuber2022annotatedhistorymodernai}. It was selected, because it's significance extends beyond mere categorization: more intricate tasks, like object detection, semantic segmentation, and image generation, are derived from the same principles~\cite{Szeliski2022}. This allows to assume a similar behavior of the mixed training strategies on a broad range of tasks in computer vision.

\subsection{Artificial Neural Networks}
As noted before, mixed training can be used irrespective of the ANN architecture. Therefor, we decided to assess the impact of three commonly used architectures in AI-based computer vision. 
Namely, the multilayer perceptron (MLP)~\cite{schmidhuber2022annotatedhistorymodernai}, the convolutional neural network (CNN)~\cite{Fukushima1980}, and the vision transformer (ViT)~\cite{DBLP:conf/iclr/DosovitskiyB0WZ21}. Each of these ANNs is implemented in the original form, without additional mechanisms, like  normalization or regularization layers. This reduces the number of confounding variables and allows to focus on the respective ANN's core mechanism. However, normalization layers were incorporated into the transformer blocks and the patch encoder of the ViT, given their critical role in preserving the fundamental functionality and performance of transformer-based networks~\cite{10.5555/3524938.3525913}. The exclusion of these layers would constitute a deviation from the fundamental design principles of transformers~\cite{NIPS2017_3f5ee243}. The general design choices, such as the width of the layers, the activation functions, the initialization, etc., reflect common practices. A detailed overview of the network configurations can be found in \ref{footnote_code}.

\subsection{Datasets}\label{datasets}
Synthetic data can be generated through several distinct mechanisms~\cite{Niko}. 
For this comprehensive study on mixed training strategies, we curated three hybrid datasets, each created using a different synthetic data generation mechanism. 
These are generative Ai (GenAI), computer aided design (CAD) and hand drawings, which results in distinct structures and characteristics of each dataset.

\begin{figure}[t]
\centering
\scalebox{0.9}{
\begin{subfigure}[b]{0.3\textwidth}
\centering
\includegraphics[height=4cm,keepaspectratio,alt={Example image from the real Cifar-10 Dataset. A red sports car is parked on a smooth surface, with a sleek and aerodynamic design. The car features prominent headlights, a low profile, and shiny alloy wheels.}]{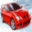}
\caption{Cifar-10 (real)}
\label{fig:fig1_a}
\end{subfigure}%
\hspace{0.04\textwidth}
\begin{subfigure}[b]{0.3\textwidth}
\centering
\includegraphics[height=4cm,keepaspectratio,alt={Example image from the real LegoBricks Dataset. A small green lego brick with a cylindrical connector on top and a hollow side attachment, placed on a light surface.}]{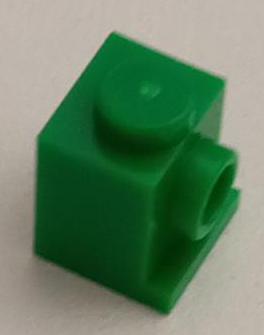}
\caption{Lego (real)}
\label{fig:fig1_b}
\end{subfigure}
\hspace{0.04\textwidth}%
\begin{subfigure}[b]{0.3\textwidth}
\centering
\includegraphics[height=4cm,keepaspectratio,alt={Example image from the real DomainNet Dataset. A wooden boomerang featuring traditional Aboriginal art. The boomerang is decorated with a painted kangaroo, geometric patterns, and dot motifs in earthy tones of brown, red, yellow, and black. The design includes two oval shapes and a circular pattern.}]{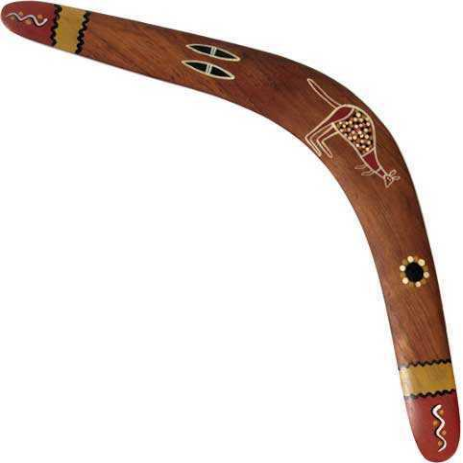}
\caption{DomainNet (real)}
\label{fig:fig1_c}
\end{subfigure}
\vspace{0.01\textwidth} 
}

\scalebox{0.9}{
\begin{subfigure}[b]{0.3\textwidth}
    \centering
    \includegraphics[height=4cm,keepaspectratio,alt={Example image from the synthetic CiFake Dataset. A classic red convertible car is parked on a street with a blurry background. The car's design features a sleek body and chrome accents.}]{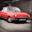}
    \caption{CiFake (synth.)}
    \label{fig:fig1_d}
\end{subfigure}%
\hspace{0.04\textwidth}%
\begin{subfigure}[b]{0.3\textwidth}
    \centering
    \includegraphics[height=4cm,keepaspectratio,alt={Example image from the synthetic LegoBricks Dataset. A 3D render of a lego brick consisting of a cube with a hollow square cutout on the top surface. Attached to one side of the cube is a cylindrical protrusion.}]{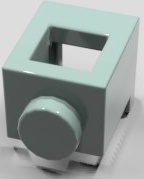}
    \caption{Lego (synth.)}
    \label{fig:fig1_e}
\end{subfigure}%
\hspace{0.04\textwidth}%
\begin{subfigure}[b]{0.3\textwidth}
    \centering
    \includegraphics[height=4cm,keepaspectratio,alt={Example image from the synthetic DomainNet Dataset. A simple black line sketch depicting an irregular, arch-like shape with a pointed end on the left and a rounded end on the right. The shape resembles a boomerang. The background is plain white.}]{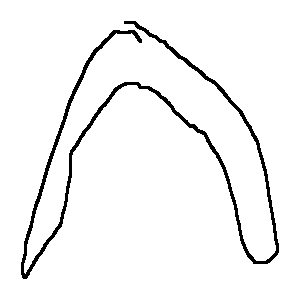}
    \caption{DomainNet (synth.)}
    \label{fig:fig1_f}
\end{subfigure}
}
\caption{Real and synthetic example images of the datasets.}
\label{fig:six_example_images}
\end{figure}

\subsubsection{Cifar-10 and CiFake} comprise the first dataset in which, as illustrated in Fig.~\ref{fig:six_example_images}~(\subref{fig:fig1_a}) and (\subref{fig:fig1_d}), Cifar-10~\cite{krizhevsky2009learning} represents the real and CiFake~\cite{cifake} the synthetic part. Cifar-10 and CiFake both comprise $60,000$ $32\times32$ RGB images across ten object classes, each with $6,000$ images. The CiFake dataset is generated using the \textit{CompVis Stable Diffusion} model~\cite{Rombach_2022_CVPR}, an open-source latent diffusion model. Additional prompt modifiers enhanced the diversity, while still producing images that closely resemble Cifar-10's characteristics. Importantly, $668$ duplicate image pairs were found in the CiFake dataset, and one from each was removed to ensure dataset integrity.

 \subsubsection{LEGO Bricks}~\cite{tboinski2021} for training classification networks was selected as the second dataset. It's real part is shown in Fig.~\ref{fig:six_example_images}~(\subref{fig:fig1_b}), while the synthetic images, shown in Fig.~\ref{fig:six_example_images}~(\subref{fig:fig1_e}) were generated using the computer-aided design (CAD) program LDraw~\cite{LDraw2020}. 
 The synthetic images introduce additional colors and angles, but still try to mimic the real images accurately. The dataset features 447 unique LEGO brick classes with a total of 50,000 real and 560,000 synthetic images. The classes are identified by the official LEGO IDs based on shape, regardless of color or decorations. All images are RGB with varying sizes, and both datasets are highly, yet differently, unbalanced, with class samples ranging from $6$ to over $600$.

\subsubsection{DomainNet}~\cite{peng2019moment} is the third dataset used in our study. In it's original form, it consists of six domains: Clipart, Infograph, Painting, Quickdraw, Real, and Sketch, with a total of $596,000$ samples across $345$ categories. We selected the Real domain as our real, and the Quickdraw domain as our synthetic data, as visualized in Fig.~\ref{fig:six_example_images}~(\subref{fig:fig1_c}) and (\subref{fig:fig1_f}). The former consists of roughly $176,000$ RGB images of varying sizes and class distributions, while the latter is composed of $500$ hand-drawn, black-and-white images per class, each with a size of $300\times300$. This provides a third, distinct hybrid dataset compared to CiFake (GenAI) and LegoBricks (CAD), because it's synthetic part is not designed to be as similar to the real part as possible.

\subsection{Dataset Creation}

To examine the effect of synthetic-to-real image ratios, we generated 11 equally sized subsamples from each of the three datasets introduced in Section~\ref{datasets}. 
For every original dataset we produced one subset made entirely of real images, one made entirely of synthetic images, and nine hybrid subsets whose composition ranges from 90 \% synthetic / 10 \% real to 10 \% synthetic / 90 \% real in 10-percentage-point steps.
To eliminate potential confounding factors, we matched the class distribution across all subsets. For the nine hybrid subsets, the same class balance was enforced separately within the real and synthetic parts. 
The 11 synthetic-to-real ratios on the three original datasets resulted in $33$ subsampled datasets for this study, with the attributes summarized in Table~\ref{fig:datasets}.

Furthermore, the images were rescaled to the same spatial dimensions, to match the ANN's fixed input sizes. No rescaling was necessary for Cifar-10 and CiFake, since all images have the shape $32\times32$. The LegoBricks images were rescaled to $256\times256$, and all DomainNet images to $300\times300$. Additionally, the grayscale images from the synthetic part of the DomainNet dataset were converted into the $3$-channel RGB format, for the same reason. 
Lastly, all images were normalized to lie within the range $[0, 1]$, in order to ensure consistency across all RGB channels and image sources.

\begin{table}[htbp]
\centering
\caption{Attribute overview of the (hybrid) datasets}
\label{fig:datasets}
\begin{tabular}{
|
>{\raggedright\arraybackslash}p{0.3\linewidth}|
>{\centering\arraybackslash}p{0.14\linewidth}|
>{\centering\arraybackslash}p{0.14\linewidth}|
>{\centering\arraybackslash}p{0.225\linewidth}|
>{\raggedright\arraybackslash}p{0.14\linewidth}|
}

\toprule
Name            & \# Images &  \# Classes & Image Dimensions & Synth. Source \\ \midrule
Cifar-10~\cite{krizhevsky2009learning}/Cifake~\cite{cifake} & $55,000$    & $10$         & $32\times32\times3$          &  GenAI \\
LegoBricks~\cite{tboinski2021}     & $20,100$    & $134$        & $256\times256\times3$        & CAD \\
DomainNet~\cite{peng2019moment}       & $30,000$    & $60$         & $300\times300\times3$        & Drawing \\ \bottomrule
\end{tabular}
\end{table}

\subsection{Mixed Training Strategies}
Two mixed training strategies are compared in this study, which we termed \emph{simple mixed} (SM) and \emph{fine-tuned} (FT). The former treats the synthetic and real part entirely equivalent. It does not not distinguish between synthetic and real data during training; data from both domains are utilized in the same manner, as if they were part of a single dataset. 
The inputs and their corresponding labels are sampled randomly from both datasets, regardless of their size or distribution. 
This provides the ANN with a broader range of examples, increasing its ability to adapt to various scenarios and improving its robustness.

Conversely, the FT strategy utilizes the real and synthetic parts sequentially. Only the synthetic part is used to train the network in the first step. This pretraining is stopped when there is no more improvement on the real evaluation set, implemented through validation-based early stopping. 
In the second step, the real part is used to retrain the network obtained after step one. The FT strategy acknowledges the differences between data types and their relation to the real test data, and adjusts the training process accordingly.

\subsection{Training Procedure}
All models were training using vanilla Stochastic Gradient Descent (SGD) with a learning rate of $0.01$ and mini-batches of $64$ samples. Datasets were split $60 \%/20 \%/20 \%$ into training, validation, and test sets. The validation and test sets are sampled exclusively from the real part of the dataset and remained unchanged. In both SM and FT training strategies, the MLP, CNN and ViT were trained for $100$ epochs, ensuring convergence. After training, the model was restored to the state where it showed the highest validation accuracy. For the FT strategy, the first training step proceeded until the validation accuracy did not improve for 10 epochs; at that moment, early stopping triggered, and the weights from the best-performing model were preserved. In the second step, the network was trained for the remaining number of epochs. All parameters remained unchanged throughout this procedure.

To summarize, we trained three neural network architectures, each with two training strategies, on 27 hybrid datasets and six non-hybrid datasets (all-real or all-synthetic). One full training cycle therefore produced 162 models trained on hybrid data and 18 models trained on non-hybrid data, yielding 180 trained networks in total. To obtain statistically robust results, we repeated this cycle ten times, resulting in 1,800 trained networks overall. At the beginning of each training cycle, all 33 datasets were resampled.

\section{Results}
The two baselines that guide the evaluation of the mixed training strategies are the purely synthetic and purely real dataset settings. While the former provides information about the quality of the synthetic data, the latter shows the performance when no domain gap exists between the training and test data. 
Table~\ref{fig:baseline_results} reports the test accuracy of the baselines, averaged over ten repetitions. 
All purely synthetic setups achieve performance above chance level. This result indicates their applicability as a proxy for the real-world, although a large gap to the performance of the purely real setting is evident, highlighting the domain gap. The domain gap can be quantified as the difference between the purely real and purely synthetic results. The goal of mixed training strategies is to minimize this gap, which would result in a comparable performance to the purely real setups.

\begin{table}[htbp]
\centering
\caption{MLP, CNN, and ViT average test accuracy without mixing (0.0=100\% real, 1.0=100\% synthetic) averaged over ten runs.}
\label{fig:baseline_results}
%\begin{tabularx}{\textwidth}{lXXXXXX}
\begin{tabularx}{\textwidth}{l*{6}{>{\centering\arraybackslash}X}}
\toprule
    & \multicolumn{2}{c}{Cifar-10/CiFake} & \multicolumn{2}{c}{LegoBricks} & \multicolumn{2}{c}{DomainNet} \\
    & Real  & Synthetic & Real & Synthetic & Real & Synthetic \\ \midrule
MLP & $0.5090$ & $0.1328$ & $\mathbf{0.6699}$ & $\mathbf{0.1423}$ & $0.2064$ & $0.0244$ \\
CNN & $\mathbf{0.6413}$ & $\mathbf{0.1430}$ & $0.5582$ & $0.0130$ & $\mathbf{0.2870}$ & $\mathbf{0.0353}$ \\
ViT & $0.5165$ & $0.1255$ & $0.6296$ & $0.0125$ & $0.2660$ & $0.0345$ \\ \bottomrule
\end{tabularx}
\end{table}

% ft better than sm
In the mixed training setting, the two training strategies, SM and FT, have resulted in divergent performances. Across all combinations of datasets, synthetic-to-real proportions, and ANN architectures, FT has outperformed SM in $635$ out of $810$ cases; when averaged over the ten repetitions, in $69$ out of $81$.

% real-to-synth ratio
The synthetic-to-real ratio of the train data strongly influences the performance of the ANNs. This influence is two-fold: it impacts performance in general and determines the magnitude of the difference between the FT and SM strategies. The general impact on the performance was observed throughout all settings. Every increase in the real proportion leads to an improved performance. This finding was expected since more real data in the training set leads to a reduction of the domain gap to the real test data. Notably, the most significant improvement was observed for the first $10\%$ increase. The improvement made through the next increments gradually decreased. In other words, even a small amount of real data in the hybrid dataset leads to a drastic improvement in performance, and the bigger the proportion of real data in the dataset, the less improvement is made by increasing the real data proportion further. 

\begin{figure}[htbp]
\centering
\includegraphics[trim={0 0.4cm 0 0.9cm},clip, width=\linewidth,alt={Box plot showing accuracy versus the proportion of synthetic data for different strategies on the cifake dataset using the vit network. The x-axis represents the proportion of synthetic data ranging from 0.0 to 1.0, and the y-axis shows accuracy from 0.0 to 0.5. Three strategies are compared: baseline (yellow), fine-tuned (green), and simple_mixed (blue). Each strategy's performance is depicted with box plots, indicating variations in accuracy as the proportion of synthetic data increases.  Accuracy generally decreases as the proportion of synthetic data increases. This decrease is stronger for the simple_mixed strategy. For all proportions, the fine-tuned strategy yields higher accuracies as compared to the simple_mixed strategy}]{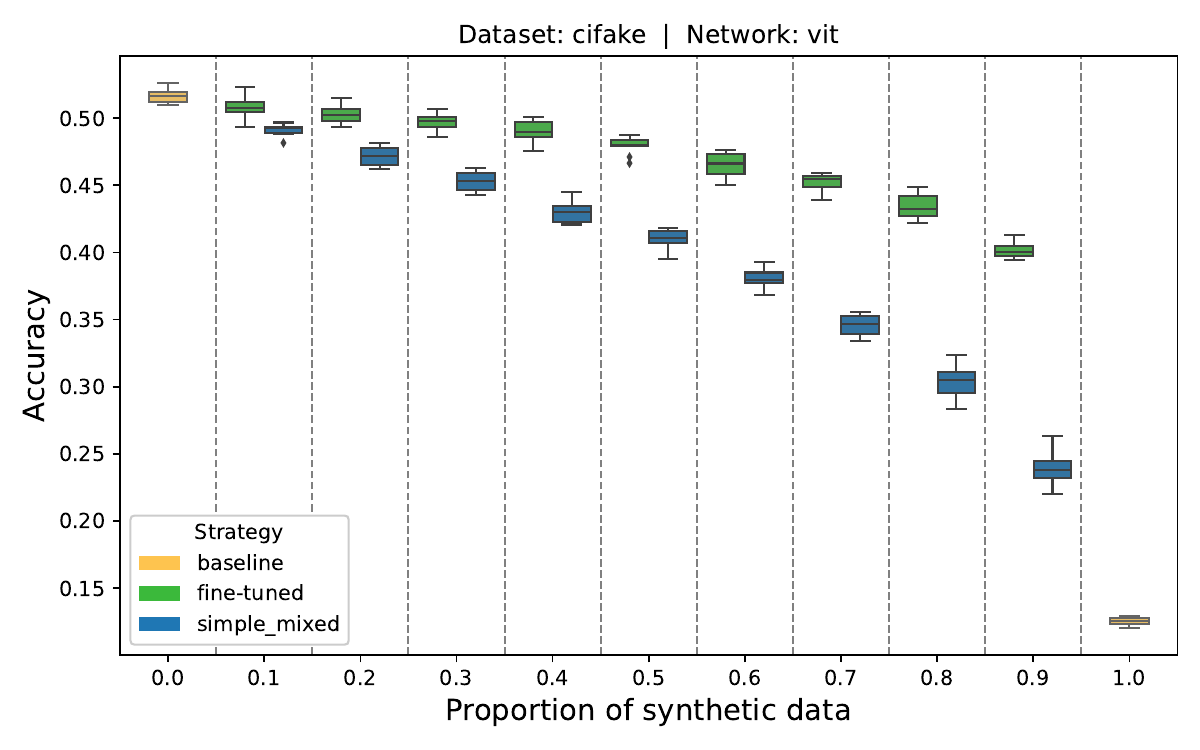}
\caption{ViT test accuracy on the Cifar-10/CiFake dataset (0.0=100\% real, 1.0=100\% synthetic) averaged over ten runs shown as boxplots.} \label{fig:cifake_vit}
\end{figure}

In addition to the general role of the synthetic-to-real ratio on the performance, it also determines the magnitude of the difference between the FT and SM strategies. Fig.~\ref{fig:cifake_vit} shows a ViT trained on the Cifar-10/Cifake dataset using both strategies. 
Here, the difference between the two strategies increases parallel to the synthetic data proportion. In the setup with only 10\% synthetic data, FT only marginally outperforms SM. With every increase of the synthetic data proportion, this difference increases as well, up to the point of 90\% synthetic data, where the FT strategy performs roughly twice as good as the SM strategy.

\begin{figure}[htbp]
\centering
\includegraphics[trim={0 0.4cm 1.1cm 1.8cm},clip, width=0.95\linewidth,alt={Line chart showing the gradient of accuracy against the proportion of synthetic data. The dataset is "cifake" and the network is "vit." Two lines are plotted: "fine-tuned" in green and "simple_mixed" in blue. Both lines show a downward trend as the proportion of synthetic data increases from 0.1 to 0.9. The y-axis ranges from -0.08 to 0.00, and the x-axis ranges from 0.1 to 0.9. The simple_mixed gradients are more negative than the fine-tuned ones, for all proportions. An increase of the downward movement can be seen for both gradient plots, but more strongly for the simple_mixed one.}]{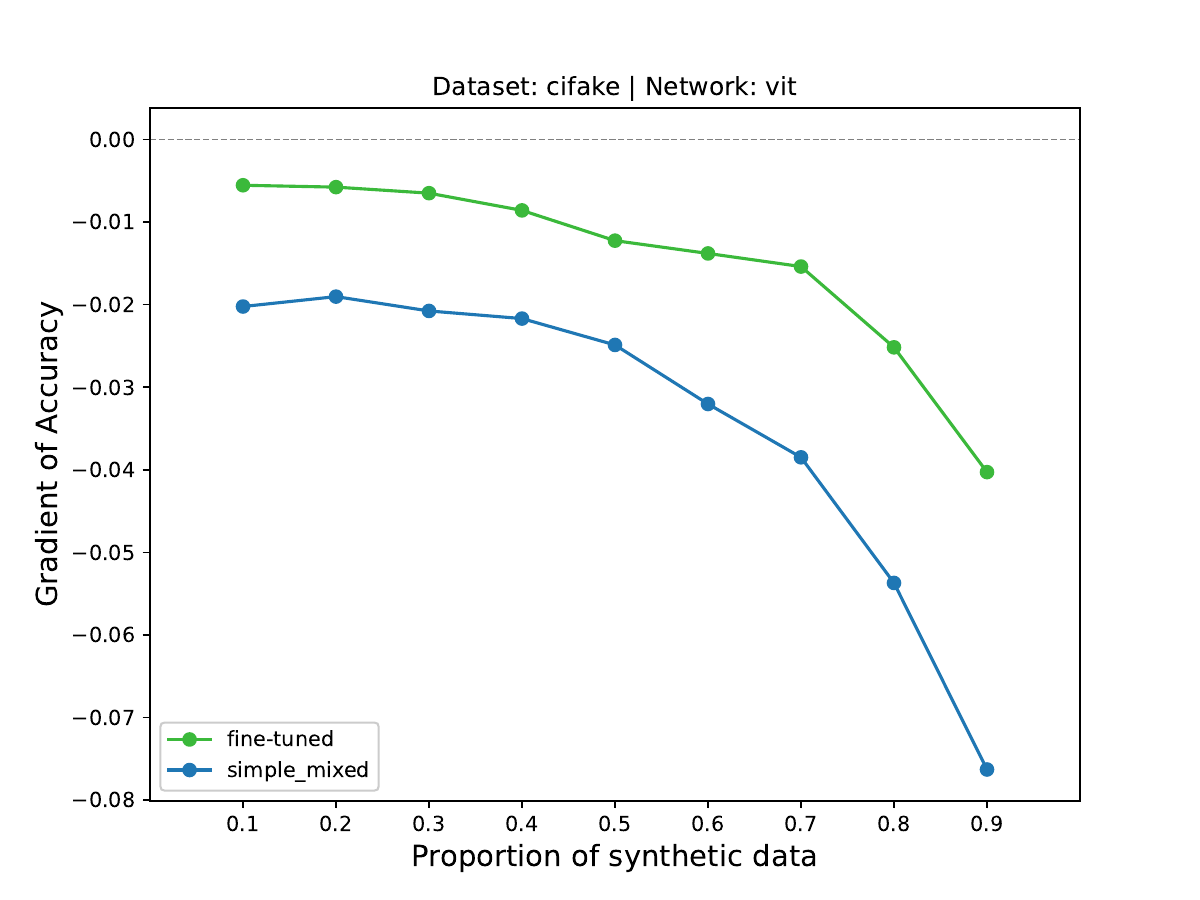}
\caption{Test accuracy gradients of the ViT trained on Cifar-10/CiFake datasets. The proportion of $0.0$ represents $100\%$ real, and $1.0$ represents $100\%$ synthetic data.} \label{fig:cifake_vit_gradient}
\end{figure}

What follows from this is that although the performance of the ANNs decrease with an increase of the synthetic proportion, this degradation of performance is less drastic for the FT strategy. 
Fig.~\ref{fig:cifake_vit_gradient} depicts the test accuracy gradients of the ViT trained on the Cifar-10/CiFake dataset with both strategies. 
The gradient shows the rate at which the performance is changing in relation to the synthetic-to-real ratio. 
The gradient of the FT strategy is constantly less negative compared to the SM strategy, highlighting the slower degradation of performance. 
In other words, the higher the proportion of the synthetic data in the dataset, the bigger is the gain in performance of the FT over the SM strategy. 
This tendency caused by the synthetic-to-real ratio was observed for all networks trained on the Cifar-10/CiFake and LegoBricks datasets, although less pronounced for the CNN.

\begin{figure}[htbp]
\centering
\includegraphics[trim={0 0.4cm 0 0.9cm},clip, width=\linewidth,alt={Box plot showing test accuracy versus the synthetic-to-real data proportion for different strategies: baseline (yellow), fine-tuned (green), and simple_mixed (blue). The x-axis represents the proportion of synthetic data ranging from 0.0 to 1.0, and the y-axis shows accuracy from 0.0 to roughly 0.225. The plot indicates a general decrease in accuracy as the proportion of synthetic data increases. For lower proportions of synthetic data, the simple_mixed strategy yields slightly higher accuracies. For higher proportions of synthetic data, the fine-tuned strategy yields slightly higher accuracies. The dataset is labeled as "domainNet" and the network as "mlp."}]{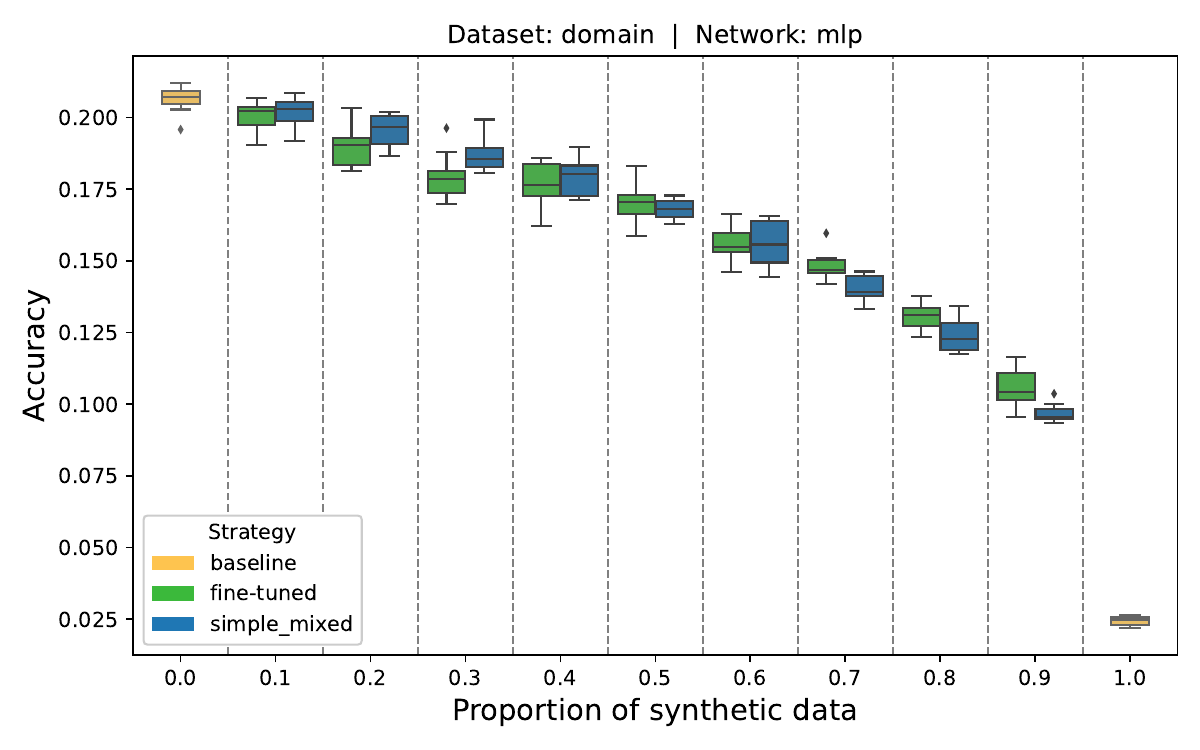}
\caption{MLP test accuracy on the DomainNet dataset (0.0=100\% real, 1.0=100\% synthetic) averaged over ten runs shown as boxplots.} \label{fig1}
\end{figure}

\begin{figure}[htbp]
\centering
\includegraphics[trim={0 0.4cm 0 0.9cm},clip, width=\linewidth,alt={Box plot showing the accuracy of different strategies—baseline (yellow), fine-tuned (green), and simple_mixed (blue)—across varying proportions of synthetic data from 0.0 to 1.0. The dataset is labeled as "domainNet" and the network as "cnn." The plot indicates a general decrease in accuracy as the proportion of synthetic data increases. Outliers are present in some data points. The lower the proportion of synthetic data, the higher is the accuracy of the simple_mixed strategy as compared to the fine-tuned strategy. With an increase of synthetic data, the difference diminishes.}]{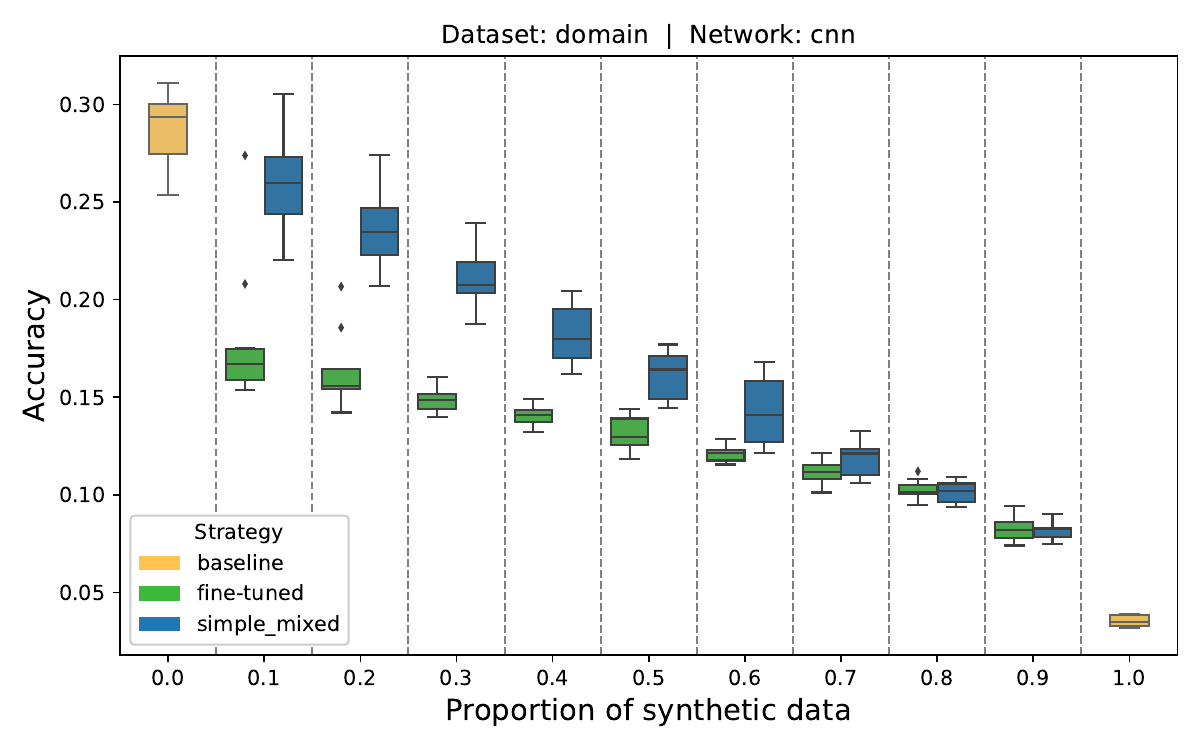}
\caption{CNN test accuracy on the DomainNet dataset (0.0=100\% real, 1.0=100\% synthetic) averaged over ten runs shown as boxplots.} \label{cnn_domain}
\end{figure}

 In the setups that included the DomainNet dataset, FT did not clearly outperform SM, as it was the case for Cifar-10/CiFake and LegoBricks. Here, the architecture of the ANN played a crucial role. For the ViT, the FT strategy consistently resulted in marginally better performances. In the setups, including the MLP shown in Fig.~\ref{fig1}, the FT strategy was better only for cases with a higher synthetic proportion. The CNN consistently performed better when trained with the SM strategy, as illustrated in Fig.~\ref{cnn_domain}. Notably, the more synthetic data is present, the smaller is the advantage of the SM strategy. 

 In addition to this preference for the SM strategy, the CNN had a higher variation throughout the 10 repetitions. The interquartile range for the SM strategy showed a high variation around the mean and the FT strategy showed significant outliers. This was observed here, as well as in the case of CNN trained on Cifar-10/Cifake and LegoBricks datasets. In Section~\ref{SupResults} of the Appendix, the results of all setups can be found.

\section{Discussion}\label{discussion}
The results of our study indicate that FT is a preferable strategy for mixed training using real and synthetic data. Nevertheless, the CNN/DomainNet setting constitutes a clear counter-example: here simultaneous exposure to both domains (SM) yielded better performance. In this section we are going to elaborate our theories for the reasons leading to this outlier.

A convolutional layer extracts localized patterns with learnable kernels; each kernel sees only its receptive field, a region whose size equals the kernel dimensions. Stacking convolutional layers (with larger kernels, strides, or pooling) expands the receptive field, such that deeper layer can capture increasingly complex structures. This, however, relies entirely on the patterns captured by earlier layers; if those first layers miss important structures, later layers have nothing to build on. High-quality, diverse features at the network’s outset are therefore critical to the overall performance.

DomainNet exhibits a larger domain gap compared to Cifar-10/CiFake and LegoBricks. The latter two contain colored, texture-matched synthetic images, meticulously designed to mimic their real counterpart, while DomainNet’s synthetic subset consists solely of black-and-white sketches with only two pixel values, sharp edges and no texture (Fig.~\ref{datasets}). 

Pre-training a CNN on these simple, synthetic images biases its early layers toward extreme black-white edges. Kernels aligned with these simple pattern receive large gradients during training, increasing a few weights rapidly, while suppressing the rest. The result is comparable to Sobel or Prewitt kernels, whose few large coefficients emphasize a single-oriented edge, whereas kernels that would encode more fine grained patterns need many smaller, finely balanced coefficients. Once dominated by such coarse detectors, the early layers fail to capture more complex patterns, leaving deeper layers without the information necessary to form complex features. The resulting kernels make subsequent fine-tuning difficult: gradients become ill-conditioned and struggle to reshape the early layers once real data are introduced in the fine-tuning step. In summary, the network is effectively trapped in a local minimum shaped by oversized and diminishing small weights.

\section{Conclusion}
Our study reveals three findings that refine the current understanding of mixed training using real and synthetic data. First, the fine-tuning strategy usually, but not always, outperforms the simple mixed strategy. Across 89 individual configurations FT surpassed SM in 69 (~78 \%), particularly when synthetic data comprised a large portion of the dataset. This indicates that it is more effective in utilizing real data to bridge the domain gap. Nevertheless, the CNN\/DomainNet setting constitutes a clear counter-example. Thus, architecture-data interactions, as outlined in \ref{discussion}, influence the strategies' success.

 Second, diminishing returns of real data. The largest performance gains occurred with an initial 10\% increase in real data; further increases yielded progressively less. Practically, adding a modest subset of real data to a synthetic set can markedly boosts performance, providing a possible cost-effective solution in many cases.

Third, Interaction of strategy with domain gap
For the two synthetic sources that explicitly mimic the real domain (GenAI and CAD), FT was consistently advantageous. Conversely, hand-drawn sketches of the DomainNet dataset constitute large discrepancy to the real images, semantically as well as visually. Under such conditions the SM strategy can outperform FT. This suggests that the “optimal” strategy could be chosen based on a quantifiable domain gap measure.

Future work should test mixed training on richer tasks, like object detection, instance or semantic segmentation, and with production‐grade networks that include modern regularization and optimization methods. Furthermore, it should develop a robust domain-gap metric that combines low-level visual statistics with high-level semantic information. A reliable score would enable the prediction of the required real-data fraction and select the 
appropriate mixing strategy, turning today’s trial-and-error process into a more systematic workflow.

\begin{credits}
\subsubsection{\ackname}
This work is part of the AgrifoodTEF-DE project. AgrifoodTEF-DE is supported by funds of the Federal Ministry of Agriculture, Food and Regional Identity (BMLEH) based on a decision of the Parliament of the Federal Republic of Germany via the Federal Office for Agriculture and Food (BLE) under the research and innovation program ‘Climate Protection in Agriculture’.

The computation of this research was done, using computing resources of the High-Performance Computing (HPC) cluster of the Osnabrück University of Applied Sciences, which were provided by the German Federal Ministry of Research, Technology and Space (BMFTR) within the HiPer4All@HSOS project.

 \subsubsection{\discintname}
 The authors have no competing interests to declare that are
 relevant to the content of this article.

\end{credits}

\vfill

% \vfill
% $ $
\pagebreak

\appendix
\section{Appendix} \label{SupResults}
\subsection{Results on Cifar-10/CiFake}
\vfill

\begin{figure}[htbp]
\centering
\includegraphics[trim={0.4cm 0.4cm 0.3cm 0.95cm},clip, width=\linewidth,alt={Box plot showing accuracy versus the proportion of synthetic data for different strategies: baseline (yellow), fine-tuned (green), and simple_mixed (blue). The dataset used is cifake, and the network is mlp. Accuracy ranges from 0.15 to 0.55 across varying proportions of synthetic data from 0.0 to 1.0. The plot includes a legend indicating the color coding for each strategy.  Accuracy generally decreases as the proportion of synthetic data increases. This decrease is stronger for the simple_mixed strategy. For all proportions, the fine-tuned strategy yields higher accuracies as compared to the simple_mixed strategy.}]{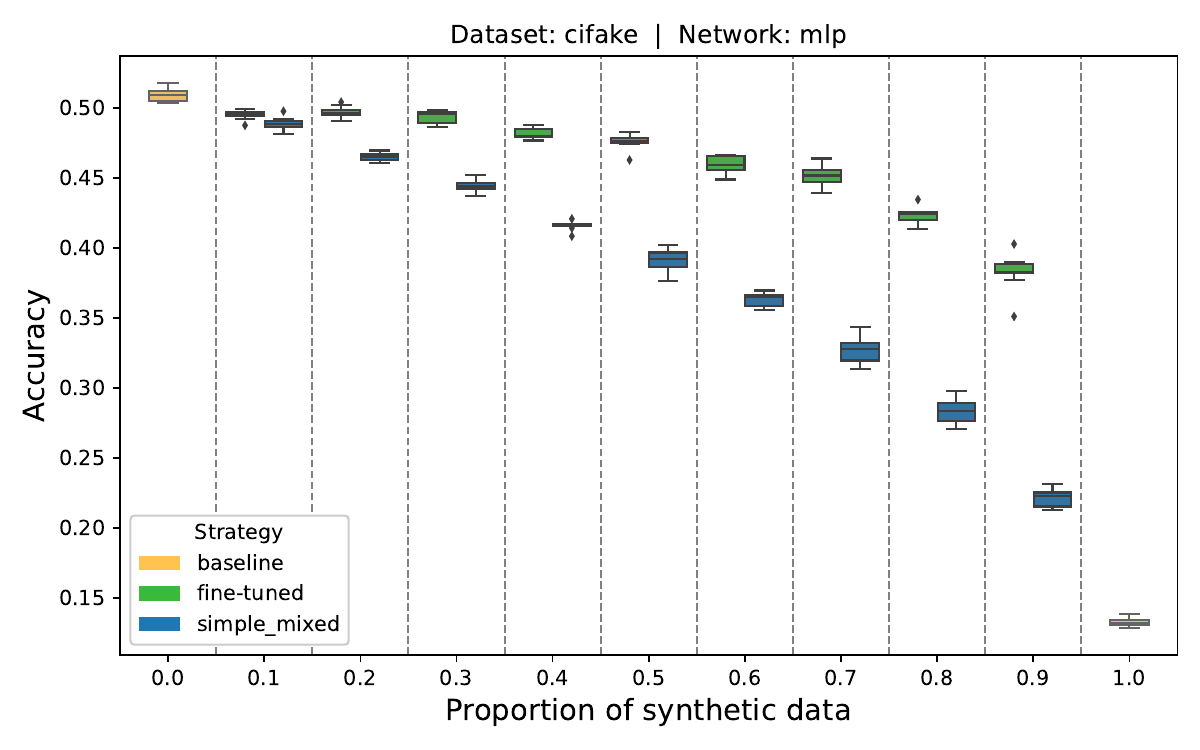}
\caption{MLP test accuracy on the Cifar-10/Cifake dataset (0.0=100\% real, 1.0=100\% synthetic) averaged over ten runs shown as boxplots.}
\label{fig:cifar_mlp}
\end{figure}
\vfill

\begin{figure}[htbp]
\centering
\includegraphics[trim={0.4cm 0.4cm 0.3cm 0.95cm},clip, width=\linewidth,alt={Box plot chart showing accuracy versus the proportion of synthetic data for different strategies: baseline (yellow), fine-tuned (green), and simple_mixed (blue).  The x-axis represents the proportion of synthetic data ranging from 0.0 to 1.0, the y-axis the accuracy. The dataset is "cifake" using a CNN network. Accuracy generally decreases as the proportion of synthetic data increases. This decrease is stronger for the simple_mixed strategy. For all proportions, the fine-tuned strategy yields higher accuracies as compared to the simple_mixed strategy. A legend indicates the color coding for each strategy.}]{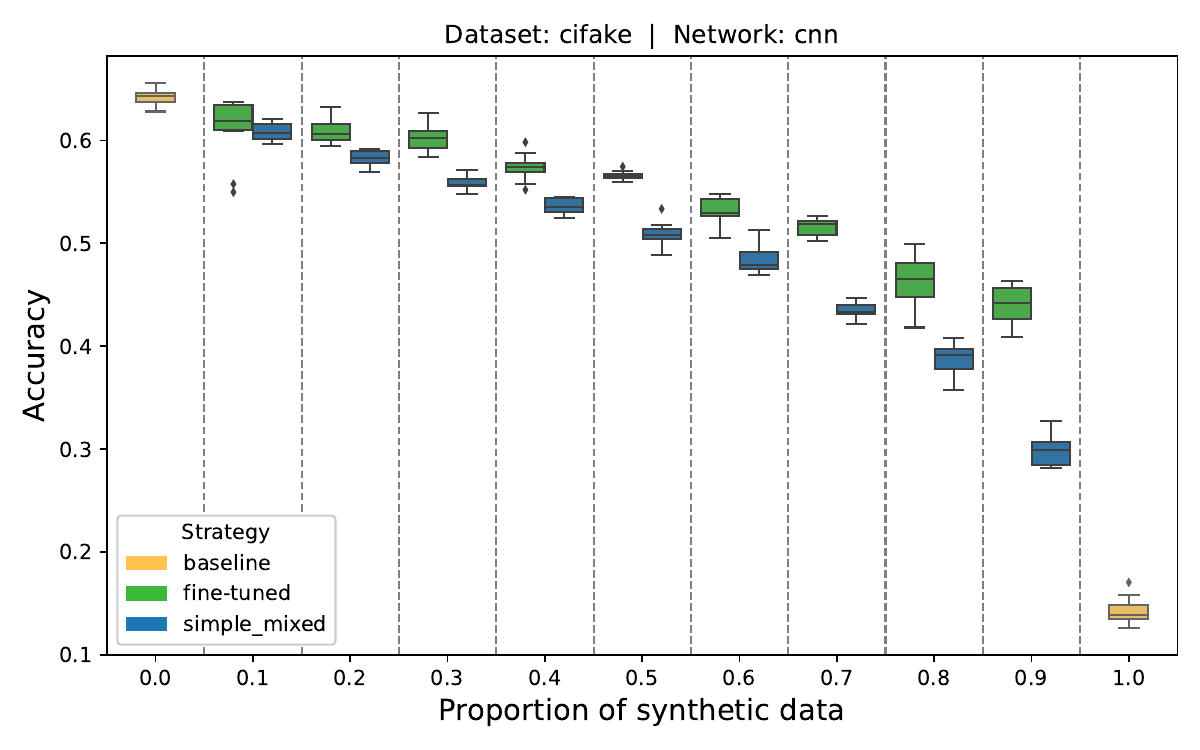}
\caption{CNN test accuracy on the Cifar-10/Cifake dataset (0.0=100\% real, 1.0=100\% synthetic) averaged over ten runs shown as boxplots.}
\label{fig:cifar_cnn}
\end{figure}

\begin{figure}[htbp]
\centering
\includegraphics[trim={0.4cm 0.4cm 0.3cm 0.95cm},clip, width=\linewidth,alt={Box plot showing accuracy versus the proportion of synthetic data for different strategies on the cifake dataset using the vit network. The x-axis represents the proportion of synthetic data ranging from 0.0 to 1.0, and the y-axis shows accuracy from 0.0 to 0.5. Three strategies are compared: baseline (yellow), fine-tuned (green), and simple_mixed (blue). Each strategy's performance is depicted with box plots, indicating variations in accuracy as the proportion of synthetic data increases.  Accuracy generally decreases as the proportion of synthetic data increases. This decrease is stronger for the simple_mixed strategy. For all proportions, the fine-tuned strategy yields higher accuracies as compared to the simple_mixed strategy}]{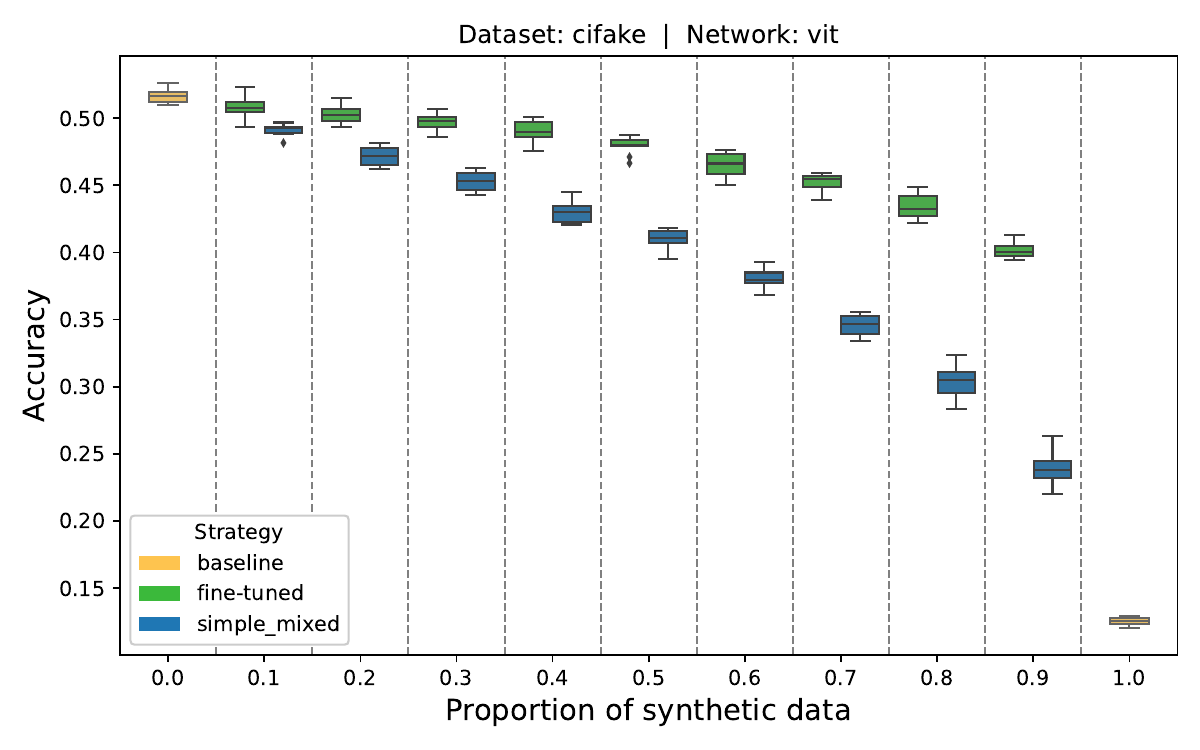}
\caption{ViT test accuracy on the Cifar-10/Cifake dataset (0.0=100\% real, 1.0=100\% synthetic) averaged over ten runs shown as boxplots.}
\label{fig:cifar_vit}
\end{figure}

\pagebreak
\subsection{Results on LegoBricks}

\vfill

\begin{figure}[htbp]
\centering
\includegraphics[trim={0.4cm 0.4cm 0.3cm 0.9cm},clip, width=\linewidth,alt={Box plot chart showing the accuracy of different strategies as a function of the proportion of synthetic data. The dataset is labeled "lego" and the network is "mlp." The strategies include "baseline" (yellow), "fine-tuned" (green), and "simple_mixed" (blue). Accuracy decreases as the proportion of synthetic data increases from 0.0 to 1.0. A legend on the left identifies the color coding for each strategy.}]{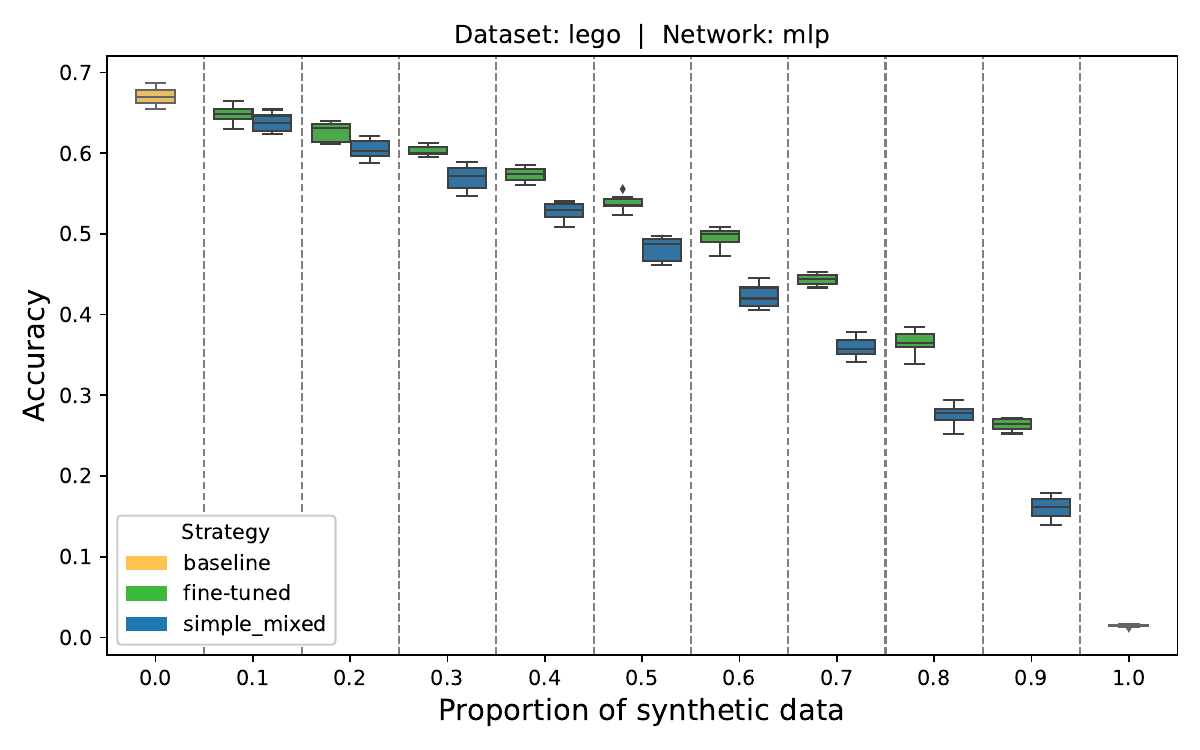}
\caption{MLP test accuracy on the LegoBricks dataset (0.0=100\% real, 1.0=100\% synthetic) averaged over ten runs shown as a boxplot.}
\label{fig:lego_mlp}
\end{figure}
\vfill

\begin{figure}[htbp]
\centering
\includegraphics[trim={0.4cm 0.4cm 0.3cm 0.9cm},clip, width=\linewidth,alt={Box plot chart showing the accuracy of different strategies as a function of the proportion of synthetic data. The dataset is labeled "lego" and the network is "mlp." The strategies include "baseline" (yellow), "fine-tuned" (green), and "simple_mixed" (blue). Accuracy decreases as the proportion of synthetic data increases from 0.0 to 1.0. A legend on the left identifies the color coding for each strategy.}]{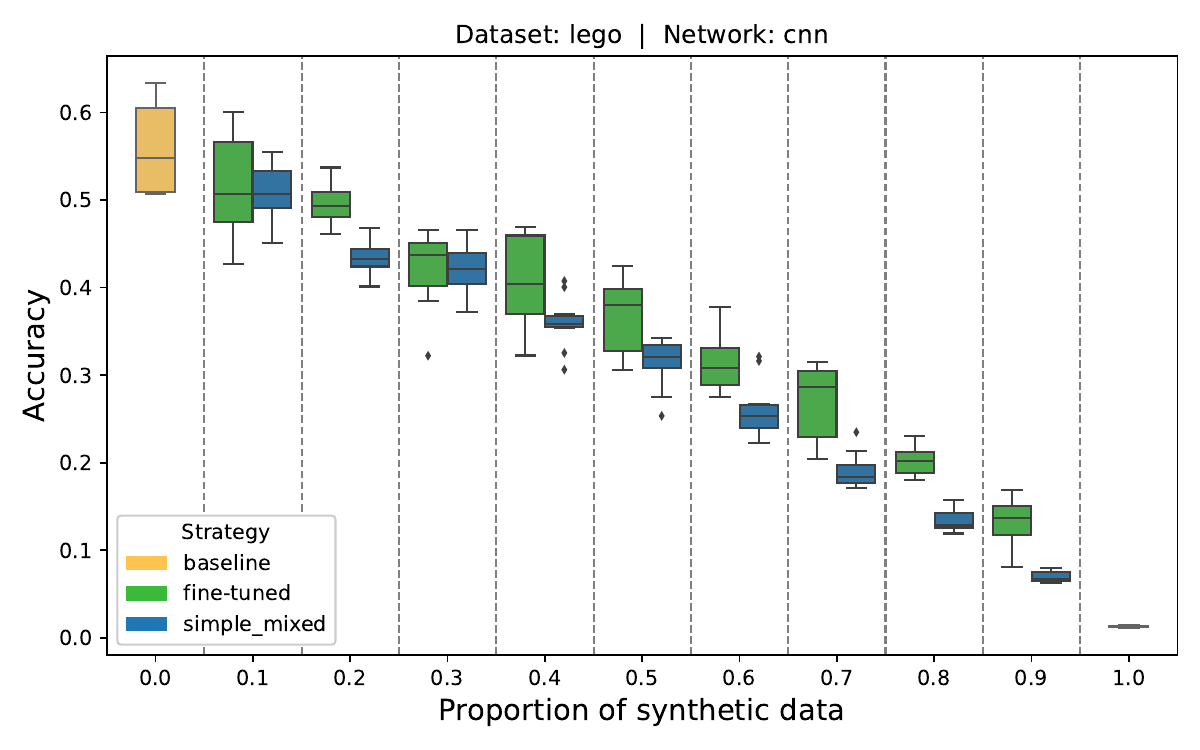}
\caption{CNN test accuracy on the LegoBricks dataset (0.0=100\% real, 1.0=100\% synthetic) averaged over ten runs shown as a boxplot.}
\label{fig:lego_cnn}
\end{figure}

\begin{figure}[htbp]
\centering
\includegraphics[trim={0.4cm 0.4cm 0.3cm 0.9cm},clip, width=\linewidth,alt={Box plot showing accuracy versus the proportion of synthetic data for different strategies: baseline (yellow), fine-tuned (green), and simple_mixed (blue). The dataset is labeled "lego" and the network is "vit." Accuracy decreases as the proportion of synthetic data increases. The plot includes a legend for strategy identification.}]{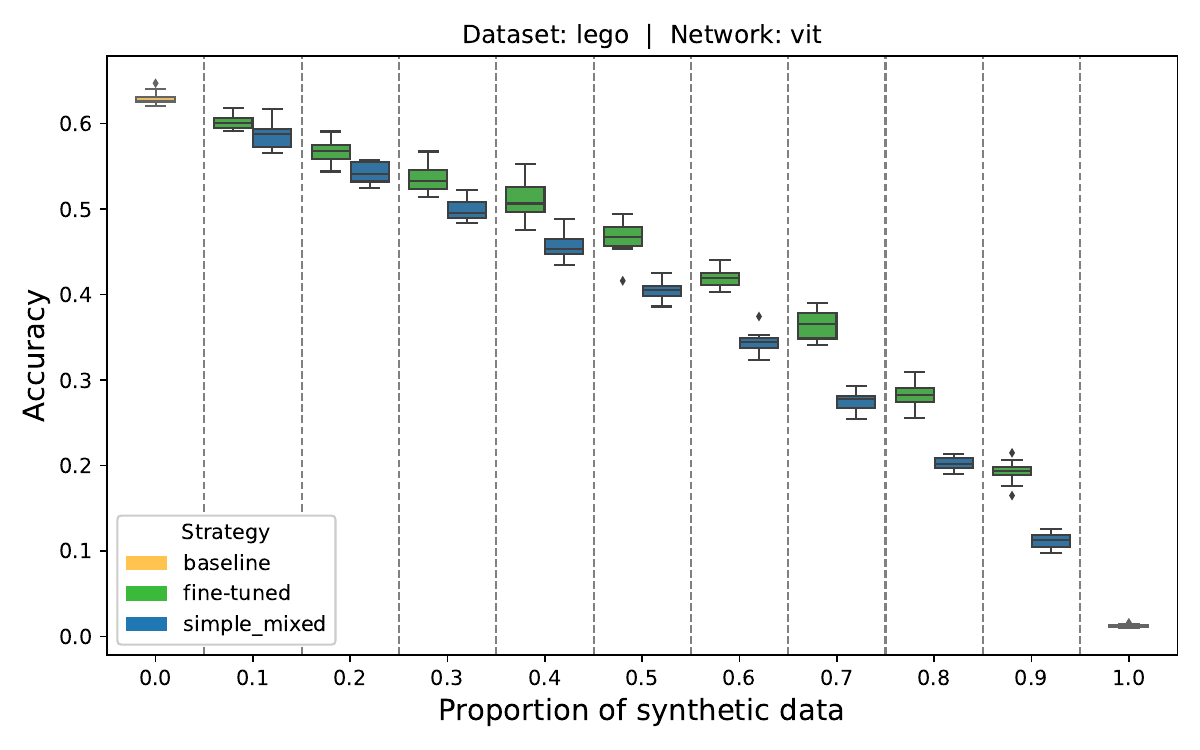}
\caption{ViT test accuracy on the LegoBricks dataset (0.0=100\% real, 1.0=100\% synthetic) averaged over ten runs shown as a boxplot.}
\label{fig:lego_vit}
\end{figure}

\pagebreak
\subsection{Results on DomainNet}
\vfill
\begin{figure}[htbp]
\centering
\includegraphics[trim={0.4cm 0.4cm 0.3cm 0.95cm},clip, width=\linewidth,alt={Box plot showing test accuracy versus the synthetic-to-real data proportion for different strategies: baseline (yellow), fine-tuned (green), and simple_mixed (blue). The x-axis represents the proportion of synthetic data ranging from 0.0 to 1.0, and the y-axis shows accuracy from 0.0 to roughly 0.225. The plot indicates a general decrease in accuracy as the proportion of synthetic data increases. For lower proportions of synthetic data, the simple_mixed strategy yields slightly higher accuracies. For higher proportions of synthetic data, the fine-tuned strategy yields slightly higher accuracies. The dataset is labeled as "domainNet" and the network as "mlp."}]{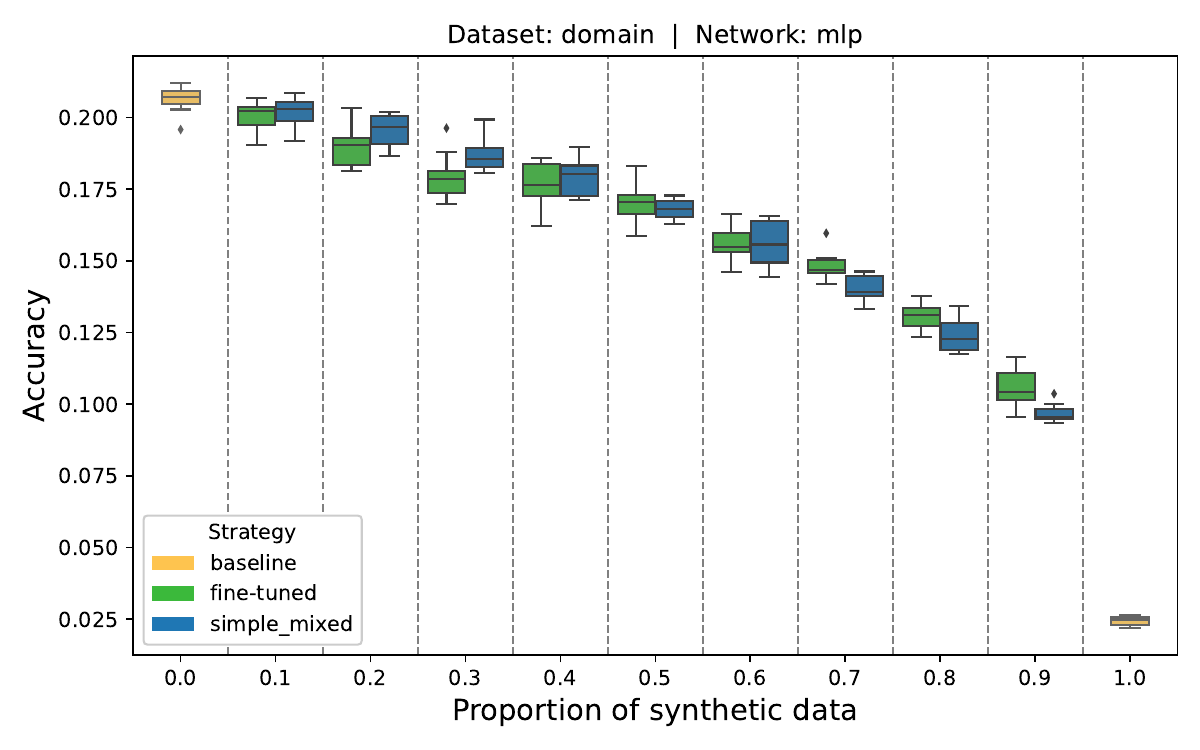}
\caption{MLP test accuracy on the DomainNet dataset (0.0=100\% real, 1.0=100\% synthetic) averaged over ten runs shown as a boxplot.}
\label{fig:domain_mlp}
\end{figure}
\vfill
\begin{figure}[htbp]
\centering
\includegraphics[trim={0.4cm 0.4cm 0.3cm 0.95cm},clip, width=\linewidth,alt={Box plot showing the accuracy of different strategies—baseline (yellow), fine-tuned (green), and simple_mixed (blue)—across varying proportions of synthetic data from 0.0 to 1.0. The dataset is labeled as "domainNet" and the network as "cnn." The plot indicates a general decrease in accuracy as the proportion of synthetic data increases. Outliers are present in some data points. The lower the proportion of synthetic data, the higher is the accuracy of the simple_mixed strategy as compared to the fine-tuned strategy. With an increase of synthetic data, the difference diminishes.}]{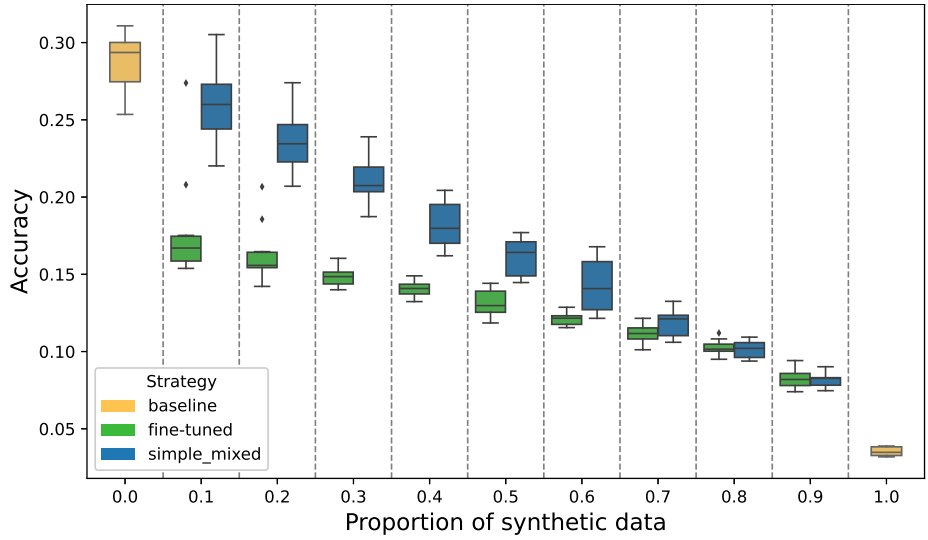}
\caption{CNN test accuracy on the DomainNet dataset (0.0=100\% real, 1.0=100\% synthetic) averaged over ten runs shown as a boxplot.}
\label{fig:domain_cnn}
\end{figure}

\begin{figure}[htbp]
\centering
\includegraphics[trim={0.4cm 0.4cm 0.3cm 0.95cm},clip, width=\linewidth,alt={Box plot showing the accuracy of different strategies—baseline (yellow), fine-tuned (green), and simple_mixed (blue)—across varying proportions of synthetic data from 0.0 to 1.0. The plot indicates a general decrease in accuracy as the proportion of synthetic data increases. he plot indicates a general decrease in accuracy as the proportion of synthetic data increases. For lower proportions of synthetic data, the simple_mixed strategy yields slightly higher accuracies. For higher proportions of synthetic data, the fine-tuned strategy yields slightly higher accuracies.  The dataset is labeled as "domain" and the network as "vit".}]{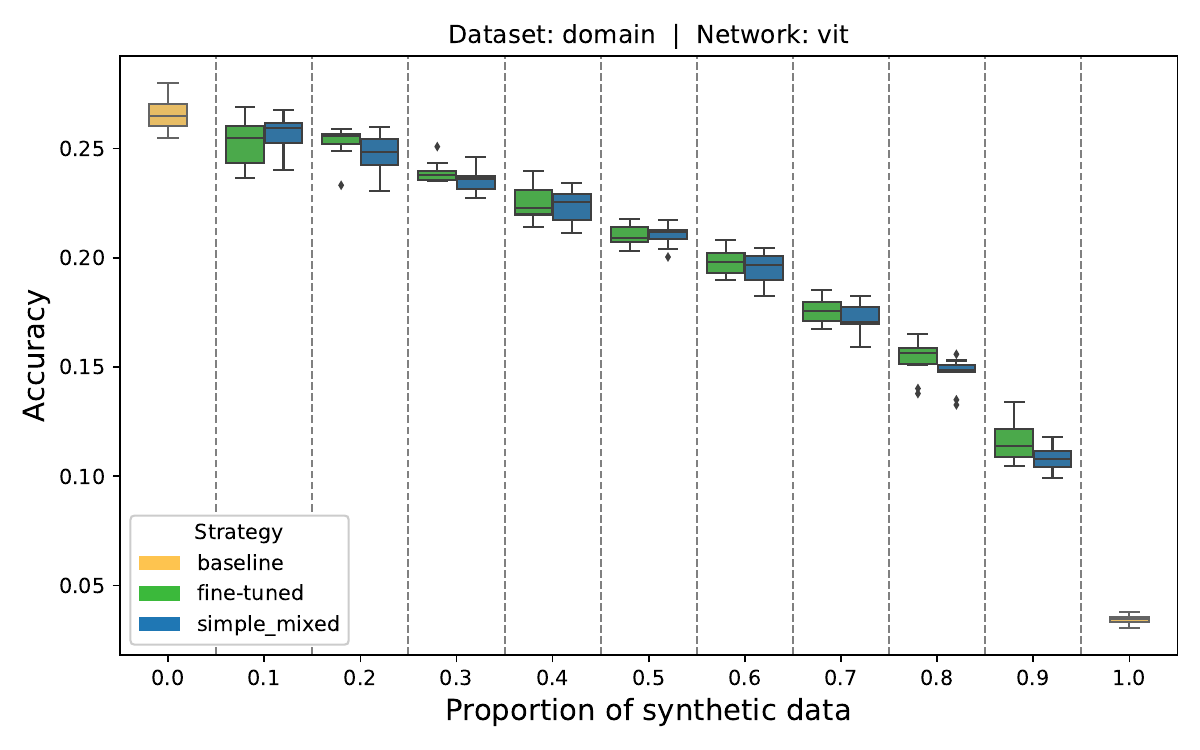}
\caption{ViT test accuracy on the DomainNet dataset (0.0=100\% real, 1.0=100\% synthetic) averaged over ten runs shown as a boxplot.}
\label{fig:domain_vit}
\end{figure}
\pagebreak

\bibliographystyle{splncs04}
\bibliography{refClean}

\end{document}